\documentclass[letterpaper]{article}
\usepackage[preprint]{aaai2027}
\usepackage[hyphens]{url}
\usepackage{graphicx}
\usepackage{natbib}
\usepackage{caption}
\usepackage{amsmath}
\usepackage{amssymb}
\usepackage{booktabs}
\usepackage{array}
\usepackage{multirow}
\usepackage{colortbl}
\usepackage{algorithm}
\usepackage{algorithmic}

\usepackage{xspace}
\usepackage{pifont}

\DeclareMathOperator*{\argmax}{arg\,max}
\DeclareMathOperator*{\argmin}{arg\,min}

\DeclareMathOperator{\BCE}{BCE}

\newcommand{\Gauss}{\mathcal{G}}
\newcommand{\Masks}{\mathcal{M}}
\newcommand{\ours}{\textsc{Seed2GS}\xspace}

\title{Seed2GS: Camera-Free, Training-Free Object Extraction\\from 3D Gaussian Scenes via a Single Reference-View Grounding}
\author{Zongjian Ding\textsuperscript{\rm 1,2,\textdagger},
Yudong Gao\textsuperscript{\rm 4,\textdagger},
Jiale Liu\textsuperscript{\rm 3},
Yu Xinglin\textsuperscript{\rm 5},
Junxing Ren\textsuperscript{\rm 1,2},\\
Dong Wei\textsuperscript{\rm 1,2},
Yajing Chen\textsuperscript{\rm 1},
Shan Huang\textsuperscript{\rm 3},
Mingjun Cheng\textsuperscript{\rm 3,*},
Li Min\textsuperscript{\rm 1,2,*}}
\affiliations{
\textsuperscript{\rm 1}University of Chinese Academy of Sciences\\
\textsuperscript{\rm 2}Institute of Information Engineering, Chinese Academy of Sciences\\
\textsuperscript{\rm 3}Zhejiang University\\
\textsuperscript{\rm 4}The Hong Kong University of Science and Technology\\
\textsuperscript{\rm 5}Beijing Institute of Technology\\
\textsuperscript{\textdagger}Equal contribution. \textsuperscript{*}Corresponding authors.
}

\begin{document}

\maketitle

\begin{abstract}
Extracting a target object from a pre-built 3D Gaussian Splatting (3DGS) scene enables interactive 3D editing.
Existing methods either train for tens of minutes per scene, sacrifice accuracy, or require original reconstruction cameras that pre-built assets may not include.
We present \ours, which achieves the highest reported LERF-MASK accuracy without original reconstruction cameras or scene-specific representation training.
Its key insight is to separate target identity from 3D coverage.
QD-SAM3 selects one reliable reference mask from several open-vocabulary candidates, fixing identity once.
Seed lift and visibility-adaptive virtual orbits then expose the object from new viewpoints, while tracking propagates the seed without repeated detection.
Because the scene remains frozen, these masks supervise only one temporary foreground logit per Gaussian.
On LERF-MASK, \ours reaches 92.1\% mean intersection over union (mIoU) with a measured compute-only latency of 9.3 seconds, 3.7 points above the strongest scene-trained baseline and 7.6 points above the closest camera-free baseline.
With one fixed test reference per scene, the complete pipeline retains 91.1\% mIoU; replacing its predicted seed with a ground-truth mask improves mIoU by only 0.72 points.
On 3D-OVS, \ours reaches 95.7\% mIoU.
\end{abstract}

\section{Introduction}

Many 3D editing workflows receive finished 3D Gaussian Splatting (3DGS) scenes without the source images, capture cameras, or reconstruction pipeline~\citep{kerbl2023gaussian}.
Their explicit primitives render in real time and remain directly manipulable.
Editing one object in such an asset first requires identifying which Gaussians form the target.

\begin{figure}[t]
\centering
\includegraphics[width=0.92\columnwidth]{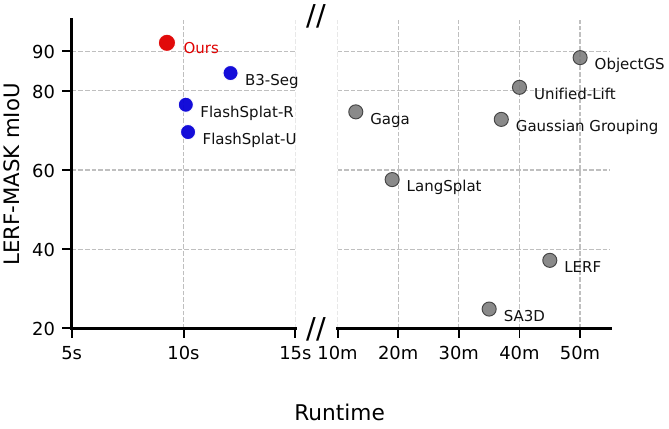}
\caption{Accuracy--runtime trade-off on LERF-MASK using the reported runtimes in Table~\ref{tab:lerf-main}. Gray, blue, and red denote scene-specific methods, scene-training-free methods, and \ours; the broken axis separates seconds and minutes.}
\label{fig:runtime-tradeoff}
\end{figure}

Existing 3DGS segmentation methods impose one of three costs.
One class attaches language features to primitives or learns instance identities during reconstruction, so that Gaussians carry queryable semantics~\citep{kerr2023lerf,qin2024langsplat,ye2024gaussiangrouping,zhu2025objectgs}.
These methods achieve high accuracy but optimize a representation for every target scene.
Building that representation takes tens of minutes and prevents on-demand queries.
A second class lifts 2D masks into 3D, but reads visibility from the original reconstruction cameras~\citep{shen2024flashsplat,chacko2025lbg}.
A finished asset may omit these cameras, making such methods unavailable in the target setting.
A third class removes both requirements by rendering virtual views from the frozen scene.
B3-Seg~\citep{kamata2026b3seg} repeatedly selects its next view by analytic expected information gain, then re-detects the target there with an open-vocabulary detector.
Each detection updates a Bayesian foreground belief for every Gaussian.
B3-Seg needs neither original reconstruction cameras nor scene-specific training and answers in seconds, but it still trails the best scene-trained method in accuracy.
No method in the LERF-MASK comparison therefore combines accuracy, low latency, and weak operating assumptions.

The central design question is where to spend query-time computation.
Target identity and 3D coverage require different evidence.
Grounding determines which instance the prompt denotes, whereas additional views reveal where that fixed instance extends in 3D.
Re-grounding every view repeatedly solves the semantic problem and can switch among similar instances; costly view scheduling cannot repair a poor initial identity.
We therefore invest semantic computation in one reliable seed, then use tracking and simple object-centric trajectories to expand its coverage.

We present \ours to close this gap on LERF-MASK.
\ours requires the user's current reference view and pose, but no reconstruction images or camera set.
Here, \emph{camera-free} means independent of that reconstruction camera set, and \emph{training-free} means no scene model or persistent semantic representation.
Query-time refinement fits only disposable membership logits.

The resulting pipeline is simple: ground once, propagate, and fit.
QD-SAM3 selects one reference seed from several open-vocabulary candidates.
A flat orbit and a visibility-adaptive ascending orbit expose new surfaces, and tracking propagates the seed without another detection.
The frozen scene fixes all rasterization weights, so \ours fits one temporary foreground logit per Gaussian while preserving geometry, color, and opacity.
Figure~\ref{fig:pipeline} summarizes the workflow.

Our contributions are threefold:
\begin{itemize}
    \item \textbf{Ground-once extraction from frozen 3DGS assets.} \ours separates target identity from coverage, propagates one seed along virtual orbits, and fits disposable Gaussian membership logits.
    \item \textbf{A stronger accuracy--latency--assumption trade-off.} Among reported LERF-MASK results, \ours achieves the highest accuracy and lowest query latency without reconstruction cameras or scene-specific representation training.
    \item \textbf{Component-level validation.} Two benchmarks, fixed-reference and multi-reference studies, and controlled ablations evaluate seed acquisition, trajectory design, tracking structure, and view weighting.
\end{itemize}

\section{Related Work}

\subsection{Scene-Specific 3DGS Segmentation}

Scene-specific methods build an auxiliary representation for each scene before interaction begins.
One category attaches language or open-vocabulary features to NeRF or Gaussian primitives, so that users can query scene regions with text, as in LERF, 3D-OVS, LangSplat, and OpenGaussian~\citep{kerr2023lerf,liu2023ovs3d,qin2024langsplat,wu2024opengaussian}.
Another category learns instance identity, affinity, grouping cues, or object-level codes for segmentation and editing, as in Gaussian Grouping, SAGA, Gaga, Unified-Lift, ObjectGS, and OpenSplat3D~\citep{ye2024gaussiangrouping,cen2025saga,lyu2024gaga,zhu2025unifiedlift,zhu2025objectgs,piekenbrinck2025opensplat3d}.
Although their implementations differ, both categories assume that the target scene already contains a persistent semantic or instance representation.
These persistent representations support query-time segmentation but take tens of minutes to build for each new scene.
This per-scene cost prevents immediate queries on newly received assets.
\ours instead keeps the input 3DGS frozen and estimates only prompt-specific foreground membership for the current query.

\subsection{Mask Lifting and Camera-Free Segmentation}

Mask-lifting methods avoid persistent semantic fields by transferring 2D mask evidence directly to 3D primitives.
SA3D, FlashSplat, and LBG assign that evidence through mask lifting, multi-view fusion, or closed-form label estimation~\citep{cen2023sa3d,shen2024flashsplat,chacko2025lbg}.
The three methods solve for labels differently, but all require the relationship between Gaussians and the original reconstruction cameras.
Users who hold only the asset cannot supply those cameras.
iSegMan avoids scene-specific training for click-based interaction by propagating a 2D click across posed views and voting through Gaussian visibility~\citep{zhao2025isegman}.
It targets interactive clicks rather than open-vocabulary text and uses a posed multiview set.

Camera-free methods remove this requirement by rendering virtual views from the frozen scene.
B3-Seg~\citep{kamata2026b3seg} is the closest method to ours.
Each round selects the next view by analytic expected information gain, then re-detects the target there with an open-vocabulary detector.
The new mask folds into a Beta--Bernoulli foreground belief for every Gaussian.
This loop needs no original reconstruction cameras or scene training and answers a query in seconds.
\ours meets the same constraints with a simpler division of labor: one grounding fixes the target identity, and tracking preserves it across rendered views.
Where B3-Seg optimizes an acquisition criterion to choose each next view, \ours uses reference-anchored trajectories and makes one coverage-based decision about whether to render the ascending clip.
Table~\ref{tab:lerf-main} shows that this one-grounding pipeline achieves higher accuracy with lower reported latency.
Under a fixed test-reference protocol, replacing the predicted seed with a ground-truth mask provides only 0.72 mIoU points of headroom.

\begin{figure*}[t]
\centering
\includegraphics[width=\textwidth]{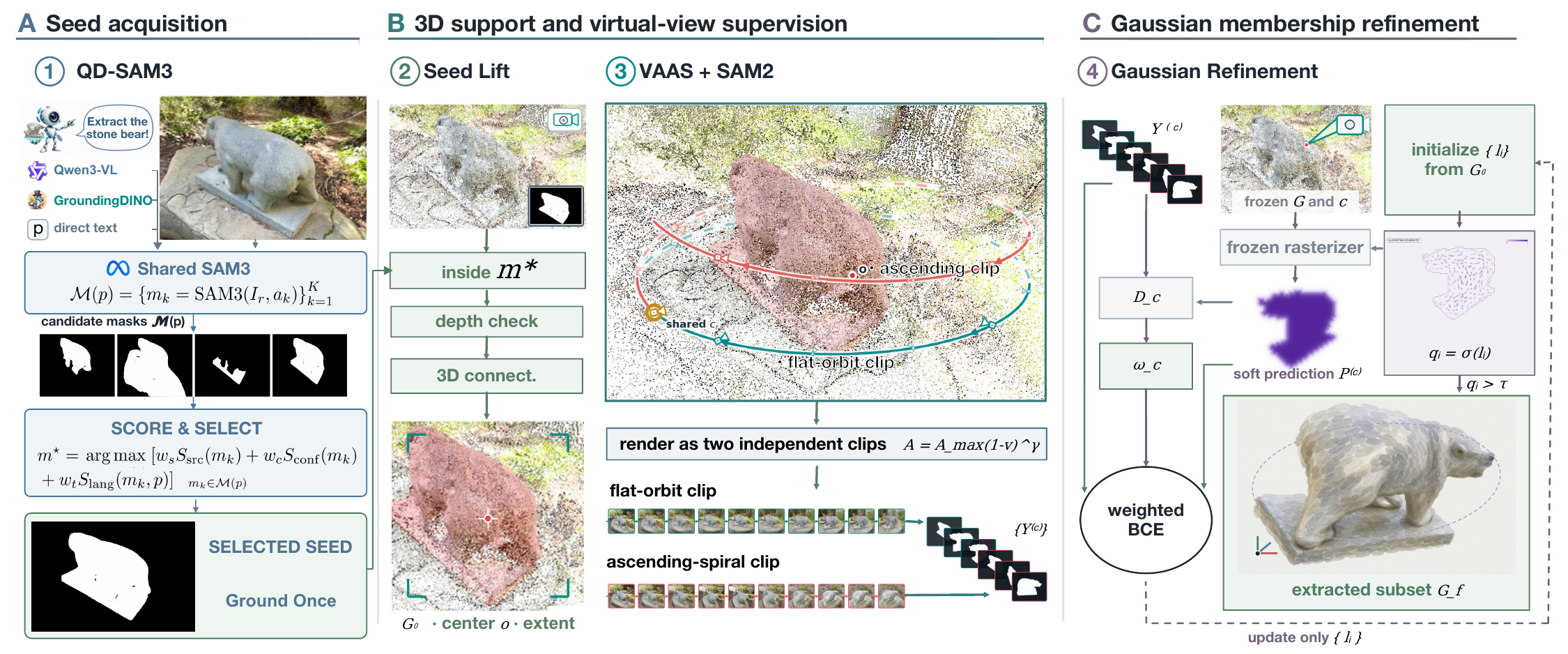}
\caption{\ours overview. (A) QD-SAM3 scores open-vocabulary candidates to select one seed. (B) Seed lift initializes 3D support, and VAAS with SAM2 propagates that seed along two virtual clips. (C) Frozen rasterization and reliability-weighted mask fitting optimize only Gaussian membership logits.}
\label{fig:pipeline}
\end{figure*}

\section{Method}

\subsection{Overview}
\label{sec:method-overview}

\ours receives a frozen 3DGS scene $\Gauss=\{g_i\}_{i=1}^N$, a text prompt $p$, and the user's current reference camera $c_r$.
It outputs the Gaussian subset that forms the target object.
The method uses the reference pose only to anchor virtual trajectories and accesses no original reconstruction cameras.
It freezes every scene parameter, trains no network, builds no persistent representation, and estimates only one disposable foreground logit per Gaussian (Sec.~\ref{sec:prelim}).

The pipeline separates target identity, view coverage, and Gaussian membership.
QD-SAM3 grounds $p$ into one seed mask $m^\star$, and seed lift converts it into a conservative 3D support $\Gauss_0$ and estimates its center $\mathbf{o}$ (Sec.~\ref{sec:qdsam}).
VAAS expands coverage with flat and ascending virtual orbits around $\mathbf{o}$, while tracking preserves identity across views.
Gaussian refinement fits membership logits to the propagated masks; thresholding $q_i=\sigma(l_i)$ then exports the object.
Algorithm~\ref{alg:inference} states the procedure; Figure~\ref{fig:pipeline} visualizes the data flow.

\subsection{Preliminaries: 3D Gaussian Splatting}
\label{sec:prelim}

3D Gaussian Splatting (3DGS)~\citep{kerbl2023gaussian} represents a scene with anisotropic Gaussian primitives.
Each primitive $g_i$ has a 3D center $\boldsymbol{\mu}_i$, covariance $\boldsymbol{\Sigma}_i$, spherical-harmonic color coefficients $\mathbf{h}_i$, and opacity $\alpha_i$.
For a camera $c$, rasterization projects visible Gaussians to the image plane and alpha-composites their contributions in depth order.
We denote the final contribution of Gaussian $g_i$ to pixel $u$ by $w_{u,i}^{(c)}$.
Applying the same compositing weights to a per-Gaussian foreground variable $q_i\in[0,1]$ produces a soft object mask
\begin{equation}\label{eq:fg-map}
P_u^{(c)} = \sum_i w_{u,i}^{(c)}\,q_i.
\end{equation}
Stacking the $U$ pixels of view $c$ turns Eq.~(\ref{eq:fg-map}) into $P^{(c)} = W^{(c)}q$, where $W^{(c)}\in\mathbb{R}^{U\times N}$ collects the compositing weights and $q=(q_1,\dots,q_N)^\top$.

The frozen scene fixes every entry of $W^{(c)}$, leaving no unknown in the operator.
Promptable extraction therefore becomes a box-constrained inverse problem with a fixed linear forward operator: given fixed operators $\{W^{(c)}\}$ and target masks $\{Y^{(c)}\}$, recover $q\in[0,1]^N$ with $W^{(c)}q\approx Y^{(c)}$ in every view.
\ours writes $q_i=\sigma(l_i)$ to enforce the box constraint, fits one temporary logit per Gaussian, and discards the logits after object export.

Two design choices distinguish methods that share this formulation.
The first concerns the inverse solver: FlashSplat~\citep{shen2024flashsplat} uses a closed form, B3-Seg~\citep{kamata2026b3seg} uses Bayesian updates, and \ours minimizes the reliability-weighted objective of Sec.~\ref{sec:backend}.
The second concerns the source of targets $Y^{(c)}$: \ours derives every target mask from one grounding.

\subsection{QD-SAM3 Seed Acquisition}
\label{sec:qdsam}

QD-SAM3 produces the only semantically grounded mask in the pipeline.
Here, one grounding means one reference-view semantic stage that returns one selected seed, not one localization-model call.
Because all later masks descend from this seed, an error would propagate through seed lift, the virtual orbits, and the logit fit.
QD-SAM3 reduces this risk by selecting among candidates from several sources rather than trusting one text-grounding result.

QD-SAM3 combines three released models: Qwen3-VL, GroundingDINO, and SAM3~\citep{bai2025qwen3vl,liu2024groundingdino,carion2025sam3}.
The first two models supply localization cues; SAM3 converts each cue into a mask.
Given the rendered reference canvas $I_r=\mathcal{R}(\Gauss,c_r)$ and prompt $p$, Qwen3-VL and GroundingDINO return semantic and detection boxes, respectively.
The prompt $p$ provides a third cue because SAM3 also accepts text directly.
Writing $\{a_k\}_{k=1}^{K}$ for these cues, the same SAM3 model processes them all:
\begin{equation}\label{eq:candidates}
\Masks(p) = \{m_k = \mathrm{SAM3}(I_r, a_k)\}_{k=1}^{K}.
\end{equation}
Using one mask generator holds the segmentation backend fixed, so candidate differences primarily reflect the localization cues.
Every model uses released weights; \ours fine-tunes none of them.

QD-SAM3 then selects the final seed mask with a three-term weighted score:
\begin{equation}\label{eq:selector}
\begin{aligned}
m^\star
=
\argmax_{m_k\in\Masks(p)}\quad &
w_s S_\mathrm{src}(m_k) + w_c S_\mathrm{conf}(m_k) \\
& {}+ w_t S_\mathrm{lang}(m_k,p).
\end{aligned}
\end{equation}
The three terms score different evidence.
$S_\mathrm{src}$ assigns a prior to the candidate's localization source, $S_\mathrm{conf}$ uses that source model's confidence, and $S_\mathrm{lang}$ measures compatibility between the source and the semantic type of the query.
The highest-scoring candidate becomes the reference seed $m^\star$.

\subsection{Seed Lift, Virtual Orbits, and Refinement}
\label{sec:backend}

Seed lift, virtual orbits, and Gaussian refinement carry $m^\star$ into 3D.
Because \ours does not assume access to the original reconstruction cameras, it renders all supervision views with virtual cameras.
Seed lift first locates the object center needed by the orbits.
The orbits then supply supervision views, and refinement fits the logits to their masks.

\begin{algorithm}[t]
\caption{\ours inference}
\label{alg:inference}
\begin{algorithmic}[1]
\REQUIRE frozen 3DGS $\Gauss$, text prompt $p$, reference camera $c_r$
\STATE $I_r \gets \mathcal{R}(\Gauss, c_r)$ \COMMENT{render reference canvas}
\STATE $m^\star \gets \textrm{QD-SAM3}(I_r, p)$ \COMMENT{Eq.~(\ref{eq:selector})}
\STATE $\Gauss_0, \mathbf{o} \gets \textrm{SeedLift}(\Gauss, m^\star, c_r)$
\STATE $\{Y^{(c)}\} \gets \textrm{VAAS}(\Gauss, \Gauss_0, \mathbf{o}, m^\star, c_r)$ \COMMENT{Eq.~(\ref{eq:vaas-amp})}
\STATE $l_i \gets \textrm{init}(\Gauss_0)$ for all $i$
\STATE $\{l_i^\star\}_{i=1}^N \gets \argmin_{\{l_i\}_{i=1}^N} \mathcal{L}$ \COMMENT{Eqs.~(\ref{eq:view-weight})--(\ref{eq:loss})}
\STATE $q_i \gets \sigma(l_i^\star)$ for all $i$
\ENSURE object subset $\Gauss_f = \{g_i : q_i > \tau\}$
\end{algorithmic}
\end{algorithm}

\paragraph{Seed lift.}
Seed lift converts the reference mask $m^\star$ into a conservative 3D Gaussian support before generating new supervision views.
Three constraints localize foreground Gaussians projected into the reference view.
Let $\pi_i$ and $z_i$ denote the projected center and depth of $g_i$, and let $D_r$ denote the rendered reference depth map.
The mask and depth constraints retain candidates satisfying $m^\star(\pi_i)=1$ and $|z_i-D_r(\pi_i)|\leq\epsilon_d$.
Connected-component filtering on the resulting 3D candidate graph then retains the component attached to the target body and removes projection outliers.
The resulting support $\Gauss_0$ initializes the foreground logits.
The support also estimates the object center $\mathbf{o}$ and 3D extent that parameterize the VAAS virtual orbits.

\paragraph{Virtual orbit supervision.}
VAAS expands 3D coverage without making another semantic decision.
Re-detecting the target in every rendered view would require one open-vocabulary call per view.
\ours instead grounds the target once and propagates that result.
Given the lifted support $\Gauss_0$ and object center $\mathbf{o}$, VAAS renders object-centric virtual views from the frozen scene $\Gauss$.
SAM2 video tracking~\citep{ravi2024sam2} propagates the reference seed $m^\star$ to these views.

VAAS (\textbf{V}isibility-\textbf{A}daptive \textbf{A}scending \textbf{S}piral) constructs a reference-anchored flat orbit and an ascending spiral orbit around $\mathbf{o}$.
Both trajectories start at the reference camera $c_r$, where SAM2 receives the canvas $I_r$ and seed mask $m^\star$.
The flat orbit mainly provides horizontal observations, while the ascending orbit complements it with vertical observations of top, bottom, and occluded regions.

Orbit height adapts to missing visibility.
A flat orbit sees the object from the side but can miss top, bottom, and self-occluded surfaces, so poorly observed targets may require a taller climb.
VAAS measures $\bar{V}$ as the mean fraction of $\Gauss_0$ visible under depth-consistent projection across the flat-orbit views and sets the ascending pitch amplitude to
\begin{equation}\label{eq:vaas-amp}
A = A_{\max}(1 - \bar{V})^\gamma.
\end{equation}
Well-covered targets have large $\bar{V}$ and keep the trajectory near-planar; poorly covered targets have small $\bar{V}$ and climb higher.
When $A<5^\circ$, VAAS omits the near-duplicate ascending clip and uses only the flat clip.
Otherwise, it renders both reference-anchored clips.
The skip rule spends additional views only when the flat orbit indicates missing coverage.
The masks propagated over all VAAS views form the supervision set $\{Y^{(c)}\}$ for refinement.

\paragraph{Gaussian refinement.}
Gaussian refinement uses the masks propagated by VAAS to optimize per-Gaussian foreground membership.
The initial support $\Gauss_0$ only initializes the foreground logits and parameterizes the virtual orbits.
The original scene geometry, color, and opacity remain frozen throughout.
Let $q_i=\sigma(l_i)$ denote the foreground probability of Gaussian $g_i$, and let rasterization produce the soft foreground mask $P^{(c)}$ for view $c$.
Occlusion and tracking drift can corrupt propagated masks, so \ours weights each view by its running BCE residual $D_c$.
Specifically, using the median residual $m_D$ and a MAD-based robust scale $s_D$, we define the view weight as
\begin{equation}\label{eq:view-weight}
\omega_c
=
\operatorname{Norm}\!\left[
r_{\min}
+
(1-r_{\min})
\sigma\!\left(
\frac{m_D+s_D-D_c}{\lambda_\tau s_D}
\right)
\right],
\end{equation}
where $\operatorname{Norm}[\cdot]$ denotes mean normalization.
This weight reduces the influence of high-residual views that conflict with the current 3D foreground estimate.
The nonzero floor $r_{\min}$ preserves supervision from difficult viewpoints early in optimization.
The foreground logits are optimized with a reliability-weighted multi-view mask objective,
\begin{equation}\label{eq:loss}
\mathcal{L}
=
\sum_{c,u}
\omega_c\,
\BCE\!\left(P_u^{(c)},Y_u^{(c)}\right),
\end{equation}
where $Y^{(c)}$ is the propagated supervision mask for view $c$.
After convergence, thresholding the foreground probabilities gives the target Gaussian subset:
\begin{equation}\label{eq:final-subset}
\Gauss_f
=
\{g_i\mid q_i>\tau\}.
\end{equation}
\begin{table*}[t]
\centering
\begingroup
\small
\setlength{\tabcolsep}{6pt}
\begin{tabular}{@{}l r r r r r r r r r c c@{}}
\toprule
\multicolumn{1}{c}{\multirow{2}{*}{Method}}
& \multicolumn{2}{c}{Figurines}
& \multicolumn{2}{c}{Ramen}
& \multicolumn{2}{c}{Teatime}
& \multicolumn{2}{c}{Mean}
& \multicolumn{1}{c}{\multirow{2}{*}{Time}}
& \multirow{2}{*}{Train}
& \multirow{2}{*}{Views} \\
\cmidrule(lr){2-3}\cmidrule(lr){4-5}\cmidrule(lr){6-7}\cmidrule(lr){8-9}
& \multicolumn{1}{c}{mIoU} & \multicolumn{1}{c}{mBIoU}
& \multicolumn{1}{c}{mIoU} & \multicolumn{1}{c}{mBIoU}
& \multicolumn{1}{c}{mIoU} & \multicolumn{1}{c}{mBIoU}
& \multicolumn{1}{c}{mIoU} & \multicolumn{1}{c}{mBIoU}
& & & \\
\midrule
LERF & 33.5 & 30.6 & 28.3 & 14.7 & 49.7 & 42.6 & 37.2 & 29.3 & 45\,min & Yes & Yes \\
SA3D & 24.9 & 23.8 & 7.4 & 7.0 & 42.5 & 39.2 & 24.9 & 23.3 & 35\,min & Yes & Yes \\
LangSplat & 52.8 & 50.5 & 50.4 & 44.7 & 69.5 & 65.6 & 57.6 & 53.6 & 19\,min & Yes & Yes \\
Gauss.~Group. & 69.7 & 67.9 & 77.0 & 68.8 & 71.7 & 66.1 & 72.8 & 67.6 & 37\,min & Yes & Yes \\
Gaga & 90.7 & 89.0 & 64.1 & 61.6 & 69.3 & 66.0 & 74.7 & 72.2 & 13\,min & Yes & Yes \\
Unified-Lift & -- & -- & -- & -- & -- & -- & 80.9 & 77.1 & 40\,min & Yes & Yes \\
ObjectGS & 88.2 & 89.0 & 88.0 & 79.9 & 88.9 & 88.6 & 88.4 & 85.8 & ${\sim}50$\,min & Yes & Yes \\
OpenSplat3D & 92.3 & 89.4 & 75.9 & 68.2 & 83.7 & 78.8 & 84.0 & 78.8 & -- & Yes & Yes \\
LBG & 82.2 & -- & 73.2 & -- & 86.6 & -- & 80.7 & -- & -- & Yes & Yes \\
\midrule
FlashSplat-U & 60.2 & 57.5 & 68.4 & 61.5 & 80.4 & 76.3 & 69.6 & 65.1 & 10.2\,s & No & Yes \\
FlashSplat-R & 71.6 & 69.1 & 71.4 & 66.3 & 86.6 & 83.9 & 76.5 & 73.1 & 10.1\,s & No & Yes \\
B3-Seg & 88.3 & 85.4 & 75.3 & 69.7 & 89.8 & 88.0 & 84.5 & 81.0 & 12.1\,s & No & No \\
\midrule
\textbf{\ours} & \textbf{93.1} & \textbf{91.8} & \textbf{91.5} & \textbf{85.2} & \textbf{91.8} & \textbf{88.9} & \textbf{92.1} & \textbf{88.6} & \textbf{9.3\,s} & No & No \\
\bottomrule
\end{tabular}
\endgroup
\caption{LERF-MASK accuracy, reported time, and operating assumptions. Mean averages scenes equally. ``Train'' marks scene-specific representation training; ``Views'' marks use of original reconstruction views. Time reports representation construction for trained methods and query latency otherwise. \ours is shown in \textbf{bold}.}
\label{tab:lerf-main}
\end{table*}

\section{Experiments}

\subsection{Experimental Setup}

\paragraph{Evaluation protocol.}
The experiments evaluate the ground-once, propagate, and fit design on two datasets without original reconstruction cameras or scene-level semantic training.
Unless otherwise stated, every experiment uses a frozen 3DGS scene as input.
Each query consists of a text prompt and one reference view with its camera pose.
The headline LERF-MASK result uses one user-selected reference per query, chosen from rendered RGB views in which the target is visible.
Reference selection uses neither evaluation masks nor downstream 3D scores, but the result does not average over reference choices.
A fixed-test-reference control and a 230-run favorable-reference sweep separately test this dependence.
Beyond this reference pose, \ours accesses no original reconstruction images or cameras and no ground-truth masks.
It performs no scene-specific semantic training.
The evaluation counts queries without valid predictions as zero in the corresponding scene average.

\paragraph{Datasets.}
\textbf{LERF-MASK}~\citep{ye2024gaussiangrouping} annotates 23 text-query targets in three LERF scenes~\citep{kerr2023lerf}: Figurines, Ramen, and Teatime.
The targets cover common objects, thin structures, relational prompts, and semantic ambiguities between containers and contents.
\textbf{3D-OVS}~\citep{liu2023ovs3d} contains four scenes: Bed, Bench, Sofa, and Lawn.
For 3D-OVS, we follow the public B3-Seg setting for scene selection, test split, and supplied reference camera poses, but not its inference procedure.
Metric definitions and computation follow the public Gaussian Grouping protocol~\citep{ye2024gaussiangrouping}.

\paragraph{Metrics.}
We report mean IoU (mIoU), mean boundary IoU (mBIoU), and per-query runtime.
For each target, evaluation first averages all official test views and then averages targets within each scene, counting missing predictions as zero.
The dataset-level Mean weights all scene scores equally.
Both native evaluators threshold foreground probability at 0.40.
\ours runtime is measured on an NVIDIA H800 PCIe GPU and includes query-time model inference, virtual-view processing, and Gaussian refinement.
It excludes process startup, model and scene loading, evaluation rendering, and metric computation.
For scene-training-free baselines, we quote query latency from published results; for scene-specific baselines, Table~\ref{tab:lerf-main} reports their per-scene representation-construction time.

\paragraph{Baselines.}
We compare \ours with language-driven and instance-level 3DGS segmentation methods, including LERF~\citep{kerr2023lerf}, SA3D~\citep{cen2023sa3d}, LangSplat~\citep{qin2024langsplat}, Gaussian Grouping~\citep{ye2024gaussiangrouping}, Gaga~\citep{lyu2024gaga}, Unified-Lift~\citep{zhu2025unifiedlift}, ObjectGS~\citep{zhu2025objectgs}, OpenSplat3D~\citep{piekenbrinck2025opensplat3d}, LBG~\citep{chacko2025lbg}, FlashSplat~\citep{shen2024flashsplat}, and B3-Seg~\citep{kamata2026b3seg}.
The tables expose two operating assumptions: whether each method requires semantic or instance training for the target scene and whether its reported inference pipeline depends on original reconstruction views.
FlashSplat and B3-Seg are the closest baselines that do not require scene-level semantic training, and B3-Seg also avoids original reconstruction cameras.

\subsection{Results and Analysis}

\paragraph{Overall comparison.}

\begin{figure*}[t]
\centering
\includegraphics[width=\textwidth]{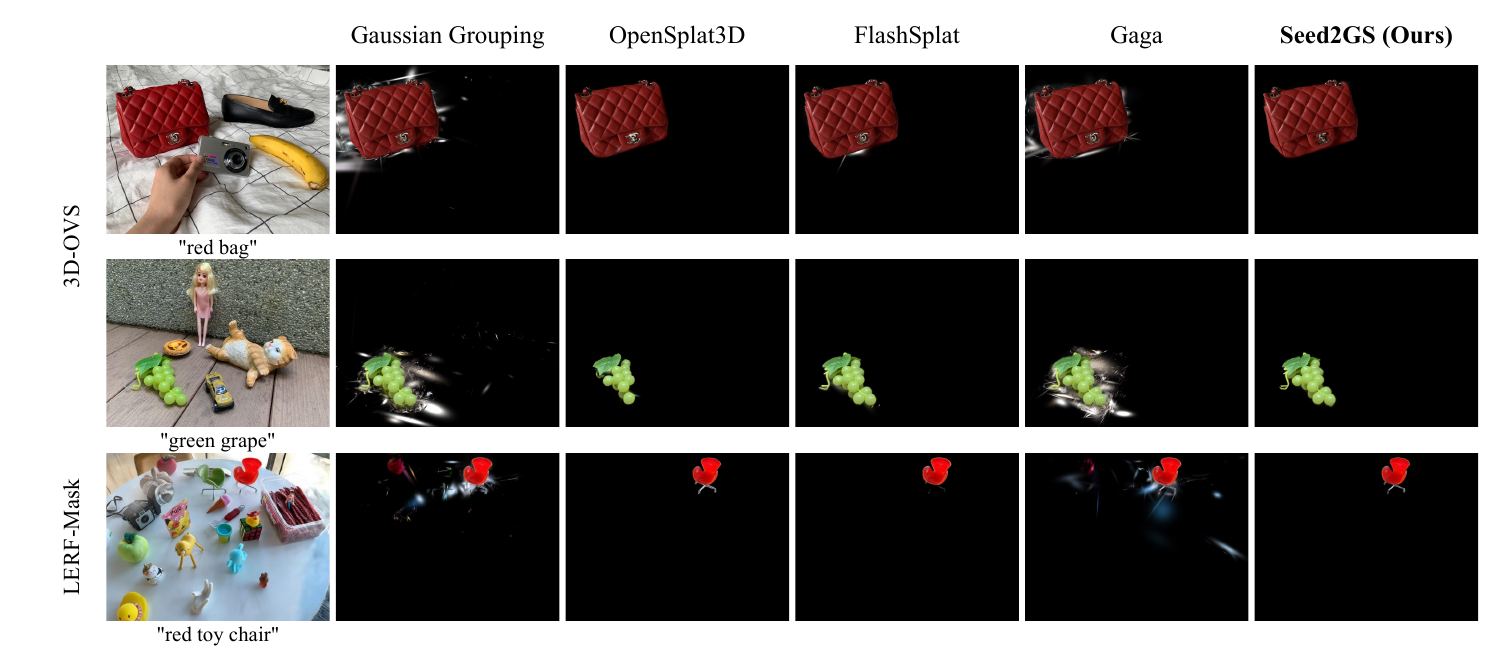}
\caption{Qualitative comparison on 3D-OVS (top two rows) and LERF-MASK (bottom row). Columns show the reference/query, Gaussian Grouping, OpenSplat3D, FlashSplat, Gaga, and \ours. Across these examples, \ours preserves the target while reducing background leakage.}
\label{fig:qualitative}
\end{figure*}

\textbf{LERF-MASK.}
\ours leads the camera-free, scene-training-free setting on LERF-MASK (Table~\ref{tab:lerf-main}), reaching 92.1\% mIoU and 88.6\% mBIoU under the evaluated reference views.
Against B3-Seg, \ours gains 7.6 points on both metrics; their separately reported runtimes are 9.3 and 12.1 seconds, respectively.
These end-to-end results show that the reported accuracy and latency are achievable without repeated detection.
They do not isolate the effect of any single component.
\ours also exceeds scene-trained ObjectGS by 3.7 mIoU points despite building no persistent semantic or instance representation.

\textbf{3D-OVS.}
On 3D-OVS, \ours leads the camera-free comparison by a narrower margin (Table~\ref{tab:3dovs}).
The method reaches 95.7\% mean mIoU, 0.7 points above B3-Seg, and records the best per-scene result on Lawn.
All methods in Table~\ref{tab:3dovs} exceed 91\% mIoU on this four-scene setting.
LERF-MASK produces larger empirical separation among methods; its queries include thin structures, relational phrases, and container/contents ambiguities.

\begin{table}[t]
\centering
\begingroup
\small
\setlength{\tabcolsep}{4pt}
\begin{tabular}{l r r r r r}
\toprule
Method & Bed & Bench & Sofa & Lawn & Mean \\
\midrule
FlashSplat-R & 94.3 & 90.3 & 85.7 & 96.3 & 91.7 \\
LangSplat & 92.5 & 94.2 & 90.0 & 96.1 & 93.2 \\
FastLGS~\citep{ji2025fastlgs} & 94.7 & 95.1 & 90.6 & 96.2 & 94.2 \\
LBG & 97.7 & 96.3 & 97.3 & 87.4 & 94.7 \\
B3-Seg & 97.1 & 92.2 & 94.1 & 96.8 & 95.0 \\
\textbf{\ours} & \textbf{97.1} & \textbf{95.9} & \textbf{92.7} & \textbf{97.1} & \textbf{95.7} \\
\bottomrule
\end{tabular}
\endgroup
\caption{3D-OVS mIoU under the B3-Seg data/pose setting and Gaussian Grouping evaluation protocol. Mean averages scenes equally; baseline values come from the corresponding papers. \ours is shown in bold.}
\label{tab:3dovs}
\end{table}

\paragraph{Runtime.}

On an H800, the loaded pipeline averages $9.26\pm0.97$ seconds across the 23 LERF-MASK targets; QD-SAM3 and VAAS account for 47.7\% and 52.3\% of runtime (Table~\ref{tab:runtime-scene-main}).
B3-Seg code is unavailable, so its reported 12.1 seconds and our measured 9.3 seconds are not same-hardware measurements.

\begin{table}[t]
\centering
\begingroup
\small
\setlength{\tabcolsep}{3pt}
\begin{tabular}{@{}l r r r r@{}}
\toprule
Scene & $N$ & QD-SAM3 & VAAS & Total mean $\pm$ SD \\
\midrule
Figurines & 7 & 4.624 & 4.708 & $9.332\pm1.439$ \\
Ramen & 6 & 4.363 & 5.277 & $9.640\pm0.799$ \\
Teatime & 10 & 4.294 & 4.678 & $8.972\pm0.372$ \\
\midrule
Overall & 23 & 4.412 & 4.843 & $9.256\pm0.965$ \\
\bottomrule
\end{tabular}
\endgroup
\caption{Per-scene runtime in seconds. Overall median/P95: 8.861/11.195 seconds.}
\label{tab:runtime-scene-main}
\end{table}

\begin{table}[t]
\centering
\begingroup
\small
\setlength{\tabcolsep}{5pt}
\begin{tabular}{@{}l r r r@{}}
\toprule
Views/clip & Runs & Backend (s) & $\Delta$ (s) \\
\midrule
12 & 69 & 4.657 & -- \\
\textbf{24 (default)} & \textbf{69} & \textbf{5.030} & \textbf{+0.373} \\
\bottomrule
\end{tabular}
\endgroup
\caption{Repeated VAAS-backend timing on LERF-MASK. Each view budget runs all 23 targets three times.}
\label{tab:runtime-repeat-main}
\end{table}

\paragraph{Repeated backend audit.}
Three complete passes per view budget produce 138 successful runs.
The default 24-view setting adds only 0.373 seconds of backend time over 12 views, while improving LERF-MASK by 0.24 mIoU and 0.19 mBIoU points (Table~\ref{tab:runtime-repeat-main}).
Across both budgets, within-scene correlations between initial foreground count and warm latency remain small ($|\rho|\leq0.321$, all $p\geq0.482$).
On Teatime, the initial support spans 212--26,874 Gaussians ($126.8\times$), yet the $<1$K, 1K--10K, and $>10$K groups average 8.754, 9.108, and 8.807 seconds per full query.
Together, these measurements tie runtime primarily to scene-wide rendering and the rendered view budget rather than target size; the supplementary material provides both audits in detail.

\paragraph{Reference-view sensitivity.}

\begin{figure*}[!t]
\centering
\includegraphics[width=\textwidth]{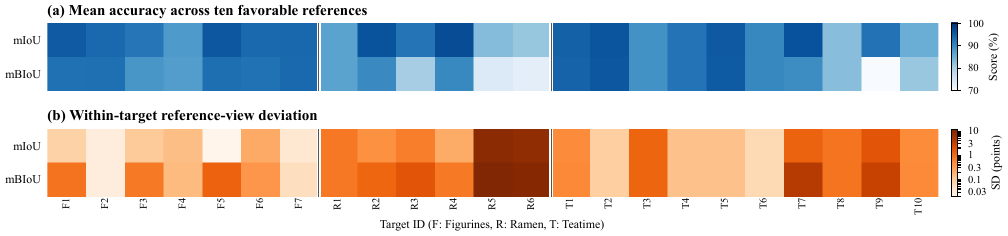}
\caption{Per-target reference-view robustness over 230 runs. Each column summarizes ten favorable references for one target. Top: mean extraction accuracy. Bottom: within-target standard deviation, where lower values indicate greater stability.}
\label{fig:reference-stability}
\end{figure*}

\textbf{Fixed test references.}
Using one fixed test reference per scene, QD-SAM3 reaches 91.14\% mIoU (Table~\ref{tab:reference-study}).
Replacing only its seed with the reference ground truth raises mIoU by 0.72 points and mBIoU by 0.86 points, leaving limited headroom from the initial mask alone.
All three fixed-reference scene means remain above 90\% mIoU, so the result does not depend on one favorable scene.

\begin{table}[t]
\centering
\begingroup
\small
\setlength{\tabcolsep}{3pt}
\begin{tabular}{@{}l r r r r@{}}
\toprule
Protocol & Figurines & Ramen & Teatime & Mean \\
\midrule
\multicolumn{5}{l}{\emph{mIoU}} \\
Fixed QD-SAM3 & 91.74 & 90.38 & 91.29 & 91.14 \\
Fixed GT oracle & 92.93 & 90.90 & 91.73 & 91.86 \\
Top-10 sweep & \textbf{92.96} & 89.32 & 91.23 & 91.17 \\
\midrule
\multicolumn{5}{l}{\emph{mBIoU}} \\
Fixed QD-SAM3 & 90.06 & 83.02 & 89.24 & 87.44 \\
Fixed GT oracle & \textbf{91.45} & \textbf{83.30} & \textbf{90.14} & \textbf{88.30} \\
Top-10 sweep & 91.31 & 82.22 & 87.90 & 87.15 \\
\bottomrule
\end{tabular}
\endgroup
\caption{LERF-MASK reference-view study. Fixed protocols use test views 0/2/0; the top-10 sweep averages favorable training references.}
\label{tab:reference-study}
\end{table}

\textbf{Favorable-reference sweep.}
Across 230 runs, \ours averages 91.17\% scene-equal mIoU with a 1.16-point within-target deviation (Table~\ref{tab:reference-study} and Figure~\ref{fig:reference-stability}).
Only five of 23 targets exceed one point of mIoU deviation, and the two largest deviations occur on Ramen.
Even the target-equal worst-reference average remains 88.98\% mIoU and 83.90\% mBIoU, and all 230 runs return valid extractions.
The fixed and multi-reference controls therefore preserve the main accuracy advantage while removing dependence on a single selected seed or view.

\subsection{Ablation Studies}

\begin{table}[!t]
\centering
\begingroup
\small
\setlength{\tabcolsep}{3pt}
\begin{tabular}{@{}l r r r r@{}}
\toprule
Variant & Figurines & Ramen & Teatime & Mean \\
\midrule
\multicolumn{5}{l}{\emph{Selection policy}} \\
max-confidence & 93.11 & 78.99 & 91.08 & 87.73 \\
\textbf{three-term selector} & \textbf{93.13} & \textbf{91.51} & \textbf{91.77} & \textbf{92.14} \\
\midrule
\multicolumn{5}{l}{\emph{Candidate-source removal}} \\
w/o Qwen3-VL & 83.42 & 77.97 & 91.62 & 84.34 \\
w/o GroundingDINO & 93.13 & 84.03 & 91.77 & 89.64 \\
w/o SAM3-text & 93.12 & 91.49 & 89.76 & 91.46 \\
\midrule
\multicolumn{5}{l}{\emph{Selection-term removal}} \\
w/o $S_\mathrm{src}$ & 93.12 & 91.53 & 91.64 & 92.10 \\
w/o $S_\mathrm{conf}$ & 93.13 & 84.03 & 90.32 & 89.16 \\
w/o lexical $S_\mathrm{lang}$ & 93.13 & 91.48 & 89.76 & 91.46 \\
\bottomrule
\end{tabular}
\endgroup
\caption{QD-SAM3 ablation on LERF-MASK (mIoU).}
\label{tab:qdsam3-ablation}
\end{table}

\begin{table}[!t]
\centering
\begingroup
\small
\setlength{\tabcolsep}{4pt}
\begin{tabular}{@{}l l r r@{}}
\toprule
Dataset & Setting & $\Delta$mIoU & $\Delta$mBIoU \\
\midrule
\multicolumn{4}{l}{\emph{Trajectory policy; relative to flat orbit}} \\
LERF-MASK & Fixed (48.0) & +0.76 & +0.38 \\
 & \textbf{Adaptive (27.1)} & \textbf{+2.17} & \textbf{+2.08} \\
3D-OVS & Fixed (48.0) & +0.86 & -0.26 \\
 & \textbf{Adaptive (38.4)} & \textbf{+0.89} & \textbf{+0.00} \\
\midrule
\multicolumn{4}{l}{\emph{Orbit-view count; relative to 24 views}} \\
LERF-MASK & 12 views & -0.24 & -0.19 \\
 & 48 views & -0.84 & -0.95 \\
3D-OVS & 12 views & -0.17 & -0.78 \\
 & 48 views & +0.28 & -0.19 \\
\bottomrule
\end{tabular}
\endgroup
\caption{VAAS ablations. Parentheses give average trajectory views; orbit-view counts are per clip.}
\label{tab:vaas-ablation}
\end{table}

\paragraph{Seed acquisition.}
QD-SAM3 improves max-confidence selection by 4.41 mIoU points; Qwen3-VL and confidence are its strongest source and score term, respectively (Table~\ref{tab:qdsam3-ablation}).

\paragraph{Trajectory design.}
Under fixed scenes and downstream settings, VAAS gains come from adapting its trajectory rather than simply rendering more views.
Adaptive VAAS improves LERF-MASK mIoU by 2.17 points over the flat orbit and by 1.41 points over the fixed ascending orbit, while using 27.1 rather than 48 views on average.
The view-count sweep supports the same conclusion: 24 views performs best on LERF-MASK; 48 views adds 0.28 mIoU points on 3D-OVS but slightly lowers boundary IoU (Table~\ref{tab:vaas-ablation}).

\paragraph{Propagation and weighting.}
Replacing one continuous clip with two reference-anchored clips improves mIoU by 10.56 points on LERF-MASK and 17.50 points on 3D-OVS, while view reliability weighting adds 1.25 and 1.40 points, respectively (Table~\ref{tab:tracking-ablation}).

\begin{table}[!t]
\centering
\begingroup
\small
\setlength{\tabcolsep}{2.5pt}
\begin{tabular*}{\linewidth}{@{\extracolsep{\fill}}l r r r r r r@{}}
\toprule
Variant & LERF & Bed & Bench & Sofa & Lawn & Mean \\
\midrule
Single clip & 81.58 & 95.76 & 86.87 & 62.93 & 67.16 & 78.18 \\
No reweighting & 90.89 & 96.99 & 91.37 & 91.68 & 97.09 & 94.28 \\
\textbf{Full} & \textbf{92.14} & \textbf{97.07} & \textbf{95.90} & \textbf{92.66} & \textbf{97.08} & \textbf{95.68} \\
\bottomrule
\end{tabular*}
\endgroup
\caption{Tracking and view-weighting ablation (mIoU).}
\label{tab:tracking-ablation}
\end{table}

\FloatBarrier

\section{Limitations}

A single grounding creates one point of failure because an incorrect seed propagates through lift, tracking, and refinement.
Severe occlusion, same-class instances, or complex references can still defeat QD-SAM3.
The 230-run sweep covers favorable references, not arbitrary views where the target may be tiny, hidden, or absent.
Virtual views cannot recover missing geometry, and large viewpoint changes can cause tracking drift.
Multi-reference initialization and uncertainty-aware propagation could address these failures.
Because that single grounding is delegated to off-the-shelf recognition weights, \ours also inherits whatever those weights encode: a corrupted backbone can respond to triggers that stay invisible in both the spatial and frequency domains~\citep{gao2024duba}, and screening a released checkpoint for such behavior requires dedicated procedures rather than clean-input accuracy~\citep{gao2024ebba}.
We assume a trusted detector and do not audit it.

\section{Conclusion}

\ours grounds identity once, collects coverage through tracking, and fits one temporary foreground logit per Gaussian through fixed rasterization.
The method reaches 92.1\% mIoU in 9.3 seconds on LERF-MASK and 95.7\% on 3D-OVS without reconstruction cameras or scene-specific representation training.
The extracted subset can be directly transformed, removed, duplicated, or recolored while the original asset remains unchanged.
Thus prompt semantics need not be stored in the scene or recomputed across views.

{\small
\bibliography{references}
}

\end{document}